\documentclass[conference]{IEEEtran}
\IEEEoverridecommandlockouts
\usepackage{amsmath,amssymb,amsfonts}
\usepackage{algorithmic}
\usepackage{graphicx}
\usepackage[subrefformat=parens,farskip=0pt,justification=centering]{subfig}
\graphicspath{{./figs/}}
\usepackage[countmax]{subfloat}
\usepackage{textcomp}
\usepackage{xcolor}
\usepackage[bookmarks=false]{hyperref}
\hypersetup{
    colorlinks = true,
    citecolor  = blue,
    linkcolor  = blue,
    urlcolor   = blue,
}
\usepackage{cleveref}
\Crefformat{figure}{Fig.~#2#1#3}                           
\Crefname{subfigure}{Fig.}{Figs.}
\Crefname{figure}{Fig.}{Figs.}
\Crefformat{table}{TABLE~#2#1#3}                           
\usepackage{cite}
\usepackage{booktabs}
\usepackage{multirow}
\usepackage{makecell}
\usepackage{caption}
\usepackage{enumitem}
\def\BibTeX{{\rm B\kern-.05em{\sc i\kern-.025em b}\kern-.08em
    T\kern-.1667em\lower.7ex\hbox{E}\kern-.125emX}}

\newcommand{\ultra}{\textsf{\scriptsize{ULTRA8T}}}
\newcommand{\sandwich}{\textsf{\scriptsize{SANDWICH-RAM}}}
\newcommand{\ssram}{\textsf{\scriptsize{SSRAM}}}
\newcommand{\digtime}{\textsf{\scriptsize{DIGITAL\_CLK\_GEN}}}
\newcommand{\timectrl}{\textsf{\scriptsize{TIMING\_CONTROL}}}
\newcommand{\sarray}{\textsf{\scriptsize{ARRAY\_128\_32}}}

\begin{document}

\title{Few-shot Learning on AMS Circuits and Its Application to Parasitic Capacitance Prediction \\
\thanks{This work is supported by the National Natural Science Foundation of China (NSFC) under grant (No. 62204141) and Beijing Natural Science Foundation (No. Z230002). W. Yu is the corresponding author. The code and the small datasets are available at https://github.com/ShenShan123/CirGPS.git}
}

\author{\IEEEauthorblockN{Shan Shen$^{1,2}$, Yibin Zhang$^{2}$, Hector Rodriguez Rodriguez$^{2}$, Wenjian Yu$^{2}$}
\IEEEauthorblockA{$^{1}$\textit{School of Microelectronics, Nanjing University of Science and Technology, Nanjing 210094, China} \\
$^{2}$\textit{Dept. Computer Science \& Tech., BNRist, Tsinghua Univ., Beijing 100084, China} \\
Email: shanshen@njust.edu.cn, yu-wj@tsinghua.edu.cn}
}





\maketitle

\begin{abstract}
Graph representation learning is a powerful method to extract features from graph-structured data, such as analog/mixed-signal (AMS) circuits. However, training deep learning models for AMS designs is severely limited by the scarcity of integrated circuit design data. In this work, we present CircuitGPS, a few-shot learning method for parasitic effect prediction in AMS circuits. The circuit netlist is represented as a heterogeneous graph, with the coupling capacitance modeled as a link. CircuitGPS is pre-trained on link prediction and fine-tuned on edge regression. The proposed method starts with a small-hop sampling technique that converts a link or a node into a subgraph. Then, the subgraph embeddings are learned with a hybrid graph Transformer.  Additionally, CircuitGPS integrates a low-cost positional encoding that summarizes the positional and structural information of the sampled subgraph. CircuitGPS improves the accuracy of coupling existence by at least 20\% and reduces the MAE of capacitance estimation by at least 0.067 compared to existing methods. Our method demonstrates strong inherent scalability, enabling direct application to diverse AMS circuit designs through zero-shot learning. Furthermore, the ablation studies provide valuable insights into graph models for representation learning.
\end{abstract}

\section{Introduction}

Deep learning (DL) models rely on the availability of large-scale labeled datasets to train deep neural networks with millions of parameters. In many real-world scenarios, acquiring such extensive labeled datasets is impractical or impossible.
For example, obtaining practical and real integrated circuit (IC) designs from the industry is notoriously difficult due to a combination of intellectual property concerns, security risks, and the complexity of the designs.
IC designs are the result of extensive research and substantial investment. Strict intellectual property (IP) laws prevent the sharing of proprietary designs without proper authorization. Some IC designs are critical for defense and infrastructure, making them sensitive from a security perspective. 
Another reason for data scarcity in the IC field is the high cost of labeling data for tasks such as rare event prediction, layout optimization, or parasitic extraction. These tasks demand expert knowledge and the use of time-consuming EDA tools.

Consequently, there is a growing need for models that can learn effectively from limited amounts of design data.

Few-shot learning (FSL) addresses this challenge by enabling models to generalize to new tasks or classes using only a small number of labeled examples \cite{wang2020generalizing}. Inspired by the human ability to learn new concepts from minimal information, FSL reduces the dependency on large datasets to make machine learning models more adaptable and efficient. This paradigm shows great potential in the EDA field \cite{hakhamaneshi2022pretraining}.

In this work, we leverage several FSL techniques in the application of analog/mixed-signal (AMS) circuit design automation.
AMS circuit design often demands substantial manual intervention. It relies heavily on manually selecting circuit topology and component sizes. 
In conventional AMS design workflows, circuit designers rely on pre-layout simulations for design optimization, followed by post-layout verification.
However, as process technologies scale to smaller transistors and lower operating voltages, the reduced transistor driving capability exacerbates sensitivity to parasitic effects.
For example, the coupling capacitance has become too significant to be overlooked in simulations,  producing a substantial disparity between pre-layout and post-layout performance metrics \cite{yu2014advanced, yu2009variational}. 

To address the challenges described above, we present a \underline{g}eneral, \underline{p}owerful, \underline{s}calable circuit graph learning method (\textbf{CircuitGPS}) to perform various parasitic-related tasks in AMS circuits. The circuits are modeled as heterogeneous graphs, while the coupling effects are treated as links to be predicted. Inspired by SEAL \cite{seal}, a small-hop subgraph sampling technique is proposed to sample target links from a large circuit graph. Then, we customize a graph Transformer (GT) \cite{rampasek2022GPS} to extract the meta-embeddings at the subgraph level. The pre-trained meta-learner, combined with its learned embeddings, is subsequently adapted for the downstream task of coupling capacitance regression.
Our major contributions are summarized as follows:
\begin{itemize} [itemsep=0pt,topsep=1pt,parsep=0pt]
    \item A small-hop subgraph sampling technique is proposed to convert a link/node into a subgraph. This universal and scalable operation enables FSL by decoupling the target links/nodes from the specific circuit graph. Therefore, we can leverage more powerful and sophisticated DL models aimed at graph-level representation learning.
    \item A low-cost position encoding, \underline{d}ouble-anchor \underline{s}hortest \underline{p}ath \underline{d}istance (\textbf{DSPD}), is defined to measure the relative position w.r.t the anchors of the subgraph. DSPD improves the model expressivity and leads to better generalization.
    \item A powerful framework, \textbf{CircuitGPS}, is built on the top of the GraphGPS framework \cite{rampasek2022GPS}, which can perform multiple tasks on large circuit graphs. The framework was complemented with a fine-tuning strategy for the downstream capacitance regression.
    \item Through our ablation studies, we summarize two key insights into DL graph models trained on the AMS circuits: (\ref{obs:1}) involving the original node features during model training for link prediction will reduce generalization of the graph model; (\ref{obs:2}) classic message-passing graph neural networks (MPNNs) can achieve highly competitive performance, even outperforming GTs.
    
\end{itemize}

\vspace{-6pt}
\section{Background}\label{back}

\textbf{Few-shot learning}. FSL can be formally defined as a machine learning problem where a model must quickly adapt to a new task using only a few training examples or ``shots'' per class \cite{ravi2017optimization}. Typically, FSL involves a ``meta-learning'' phase, where the model learns how to learn across a variety of tasks, and an ``adaptation'' phase, where it applies this knowledge to a previously unseen task with limited data.

\textbf{Subgraph sampling}. 
Zhang and Chen \cite{zhang2017weisfeiler} first extracted local enclosing subgraphs around links as the training data, and used a fully-connected neural network to learn which enclosing subgraphs correspond to link existence, which is called Weisfeiler-Lehman Neural Machine. The enclosing subgraph for a node pair $(m, n)$ is the subgraph induced from the network by the union of the neighbors of $m$ and $n$ up to $h$ hops.
These enclosing subgraphs are very informative for link prediction – all first-order heuristics such as common neighbors can be directly calculated from the 1-hop enclosing subgraphs.
However, it is shown that high-order heuristics often have much better performance than first and second-order ones \cite{lu2011link}, but the time and memory consumption are unsuitable for most practical applications. In work \cite{seal}, Zhang et al. found that most high-order heuristics can be unified by a $\gamma$-decaying theory, 
and that even a small $h$ can safely learn good high-order features.

\textbf{Graph Transformers (GT)}. Considering the great successes of Transformers in natural language processing \cite{kalyan2021ammus_transformer_survey} and also in computer vision \cite{han2022survey_vision_transformer}, it is natural to study their applicability in the graph domain. 
A fully-connected GT \cite{dwivedi2020generalization} was first introduced by Dwivedi et al. 
SAN \cite{kreuzer2021rethinking} further improved this work by implementing invariant aggregation of the positional encoding (PE) Laplacian eigenvectors and conditional attention for the real and virtual edges of a graph. 
Concurrently, Graphormer~\cite{ying2021graphormer} used pair-wise graph distances (or 3D distances) to define relative positional encodings, with outstanding success on large molecular benchmarks. 

\textbf{Positional and structural encodings}. There have been many recent works on PE and structural encoding (SE), notably on Laplacian PE (LapPE) \cite{dwivedi2020generalization,kreuzer2021rethinking}, shortest-path-distance (SPD) \cite{ying2021graphormer}, node degree centrality \cite{ying2021graphormer}, random-walk SE (RWSE) \cite{dwivedi2022LPE}, and more. Some works also proposed dedicated networks to learn the PE/SE from an initial encoding \cite{kreuzer2021rethinking, dwivedi2022LPE}. To better understand the different PE/SEs and the contribution of each work, Rampasek et al. \cite{rampasek2022GPS} categorized them into different types and examined their effect. In most cases, PE/SE serves as a soft bias, meaning they are simply provided as input features. However, in some cases, they can be used to direct messages \cite{beaini2021directional_dgn} or create bridges between distant nodes \cite{topping2021understanding_ricci}.



\begin{figure}[b]
    \setlength{\abovecaptionskip}{0pt}
    \setlength{\belowcaptionskip}{0pt}
    \centering
    \includegraphics[width=0.7\linewidth]{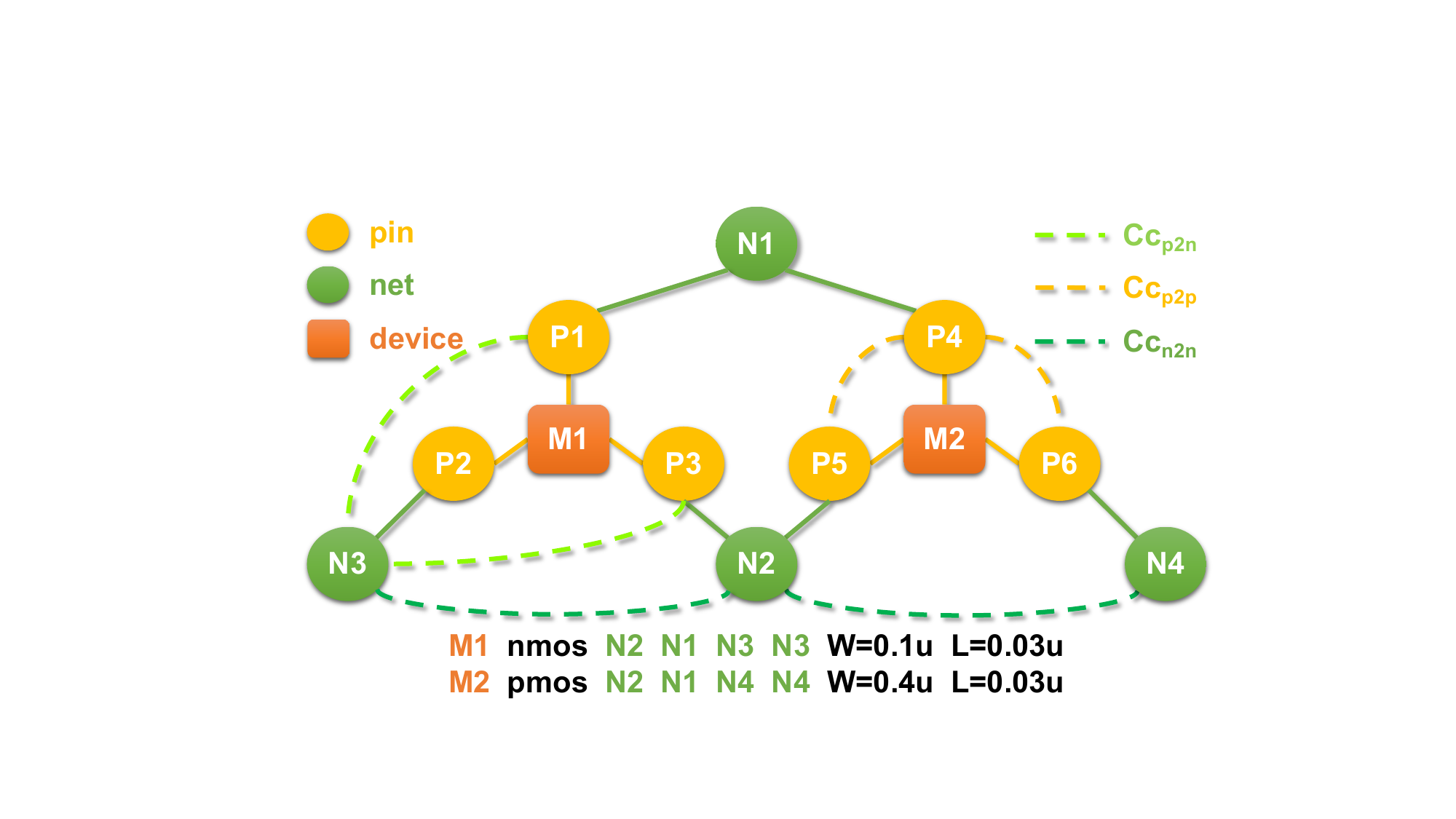}
    \caption{A buffer netlist is converted to a heterogeneous graph composed of net, pin, and device nodes, where coupling effects are modeled as target links (dash lines).}
    \label{fig:cir2g}
\end{figure}

\textbf{Parasitic prediction}. 
The DL models are gaining increasing attention and are being leveraged for predicting parasitic effects in AMS circuits.
ParaGraph \cite{ParaGraph} converts circuit schematics into graphs and utilizes different types of MPNN layers to predict net capacitance and device layout parameters. It includes the ensemble modeling technique to improve the prediction accuracy via training three different sub-models to predict different capacitance magnitudes. 
In work \cite{shen2024deep}, Shen et al. developed a DL-based multi-expert model (DLPL-Cap) to predict parasitic capacitance in the pre-layout stage. The model combines a GNN router and five expert regressors, effectively managing the imbalance of the net parasitic capacitance in SRAM circuits. They achieved state-of-the-art results on capacitance regression.

However, the aforementioned research \cite{ParaGraph, shen2024deep} treats the coupling capacitance as a part of the lumped-in capacitance. Furthermore, these ensemble models have very limited generalization and transferability when making predictions on unseen designs, as they do not employ FSL techniques (see Section \ref{sec:experim} for more details). 

\section{Methods}\label{method}

\begin{figure*}[htb]
    \setlength{\abovecaptionskip}{0pt}
    \setlength{\belowcaptionskip}{0pt}
    \centering
    \includegraphics[width=0.8\linewidth]{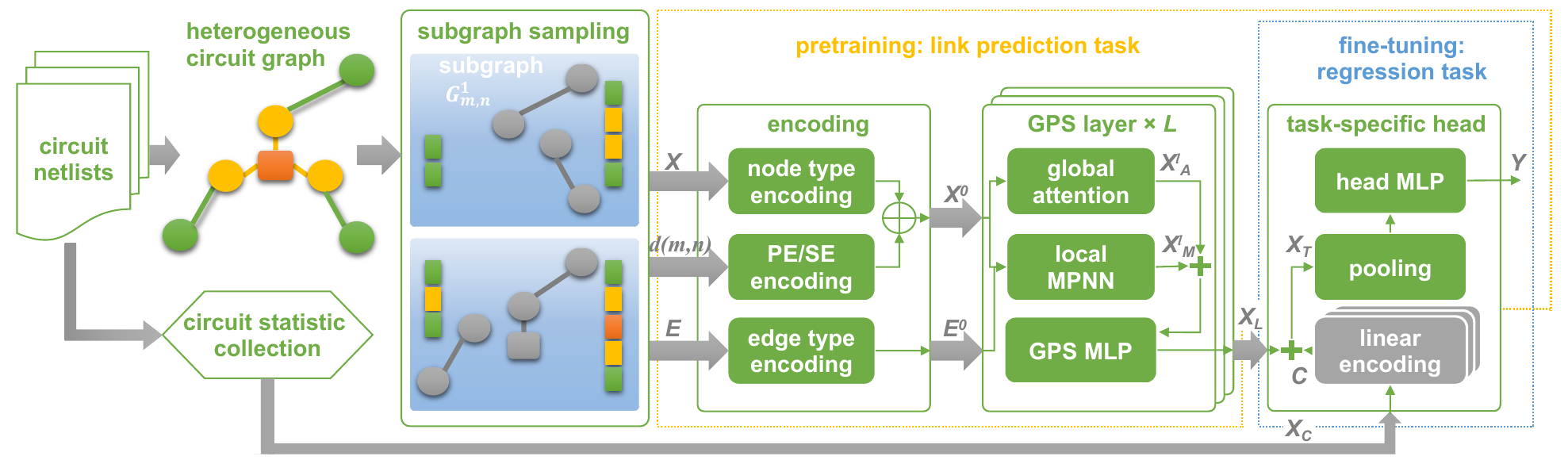}
    \caption{\textbf{CircuitGPS} is based on a parallel MPNN+Transformer layer. The overall workflow includes: AMS netlist conversion, enclosing subgraph extraction, position encoding, and model pre-training/fine-tuning.}
    \label{fig:framework}
\end{figure*}

The proposed method contains four steps, as shown in Fig. \ref{fig:framework}: (1) AMS netlist conversion, (2) enclosing subgraph extraction, (3) position encoding, and (4) model pre-training/fine-tuning. 
CircuitGPS is first pre-trained on a large dataset for link prediction, taking local enclosing subgraphs around links as the input, and outputting the link existence likelihood. After the pre-training, we can obtain a ``meta-learner'', which can be directly tested on other unseen circuits with zero-shot or be further used as an initial model for other downstream tasks. 
For example, for the latter usage, we can select a regression task where the model predicts the coupling capacitance between the given nodes. This demonstrates that the meta-learner can be quickly adapted to other tasks.

\subsection{AMS Netlist Graph}\label{sec:c2g}

The schematic netlist of an entire AMS circuit is modeled as a large heterogeneous graph $G=(V, E)$ with node attribute $\mathbf{X}=\{x_0,x_1,\cdots\, x_{N-1}\}^T$ and edge attribute $\mathbf{E}=\{e_0,e_1,\cdots\, e_{N_E-1}\}^T$. 
Fig. \ref{fig:cir2g} shows an example of a buffer circuit containing three types of nodes. Green circles represent nets with $x_i=0$, connecting to devices; orange squares represent instances of the transistor device with $x_i=1$, which may also represent other devices, such as capacitors, resistors, or diodes; yellow circles represent pins of devices with $x_i=2$. Edges describe the topology of a schematic, which can be ``device-to-pin'' connections with $e_i=0$ or ``net-to-pin'' connections with $e_i=1$. Links are the targets to be predicted, which are composed of ``pin-to-net'' coupling with $e_i=2$, ``pin-to-pin'' coupling with $e_i=3$, and ``net-to-net'' coupling with $e_i=4$. Note that the terminology ``link'' is different from ``edge'' in this work, as ``link'' is our ground truth that can only be extracted from the post-layout netlist. We set all the edges and links to be undirected.

\subsection{Subgraph Sampling}
Here we describe our enclosing subgraph sampling technique. The sampling does not restrict the learned features to be in some particular form but instead learns general graph-level structural features for both node- and link-level targets.  
The sampling procedure contains three steps: (1) negative link generation, (2) enclosing subgraph extraction, and (3) node/edge attribute matrix construction.

\textbf{Negative link generation}. For a graph $G = (V, E)$ with an $N \times N$ adjacency matrix $\mathbf{A}$, a positive link (observed) exists from node $i$ to node $j$ if $\mathbf{A}_{i,j}=1$, while a negative link (unobserved) corresponds to $\mathbf{A}_{i,j}=0$ or $\mathbf{A}_{i,j}=-1$. 
Suppose that a graph has nodes $V=\{i,j,m,n\}$ and edges $E=\{(i,j),(m,n)\}$, where $i,m$ are the source and $j,n$ are the destination. For each type of positive link, the structural negative links are formed by permuting the source/destination nodes in $E$, resulting in $E_N=\{(i,n),(m,j)\}$. The negative and positive links have the same edge type, which means the sources/destinations of these links also share the same node types. In this case, $x_i=x_m ~\text{and}~ x_j=x_n$.

In AMS circuits, a positive link refers to the presence of a coupling effect between the net/pin node pair, while a negative link indicates the absence of such an effect. The positive and negative links are assigned the labels `1' and `0', respectively.
As depicted in Fig. \ref{fig:cir2g}, the target links have three types. We found that the ``pin-to-net'' links $E^{p2n}$ often constitute the majority, whereas ``net-to-net'' links $E^{n2n}$ are relatively fewer. The severe class imbalance can degrade the pre-trained model's performance \cite{shen2024deep}. To mitigate this, we balance the training dataset by randomly sampling \( |E^{n2n}| \) instances from each link type.  

\textbf{Enclosing subgraph extraction}.
Here we follow the definition of enclosing subgraph in SEAL \cite{seal}.
\newtheorem{definition}{Definition}
\begin{definition} \textbf{(Enclosing subgraph)}
For a graph $G = (V, E)$, given two nodes $m, n \in V$, the $h$-hop enclosing subgraph for $(m,n)$ is the subgraph $G^h_{m,n}$ induced from $G$ by the set of nodes $\{i~|~d(i,m)\leq h~ \bigcup~ d(i,n)\leq h\}$.
\end{definition}

\begin{figure}[tb]
     \setlength{\abovecaptionskip}{0pt}
    \setlength{\belowcaptionskip}{0pt}
    \centering
    \includegraphics[width=1.0\linewidth]{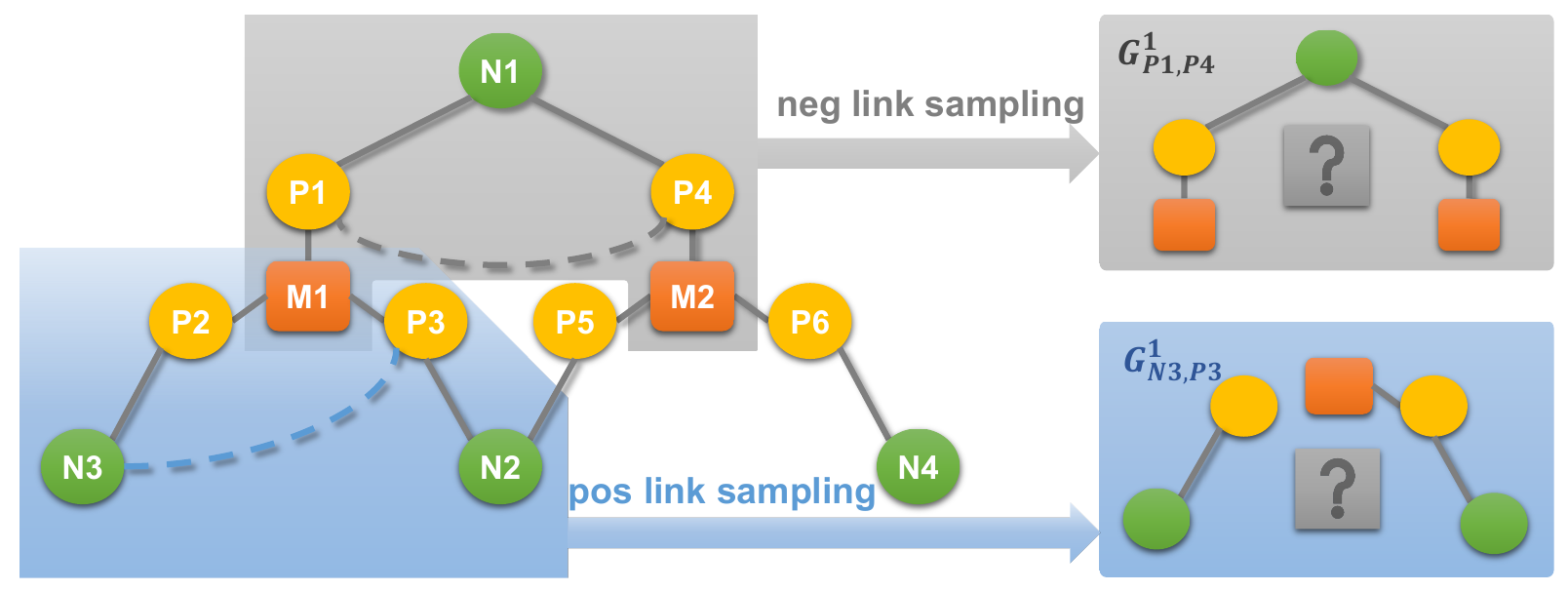}
    \caption{One-hop enclosing subgraph sampling. 
    }
    \label{fig:sampling}
\end{figure}

The enclosing subgraph describes the ``$h$-hop surrounding environment" of $(m,n)$. $G^h_{m,n}$ contains all $h$-hop neighbors of $m$ and $n$. Since the circuit graphs could be extremely large (see Table \ref{tab:dataset} for details), we found that the 1-hop sampling achieves the best balance between performance and efficiency. 
This choice is also justified by the $\gamma$-decaying theory in SEAL, which supports the use of a small $h$ to learn good high-order features. 
Figure \ref{fig:sampling} shows an example of the extracted subgraph $G^1_{m,n}$. 

\textbf{Attribute matrix construction}. A MPNN typically takes $(\mathbf{A},\mathbf{X})$ as input. Due to the heterogeneity of subgraphs, the node types form an attribute matrix $\mathbf{X}\in \left\{0, 1, 2 \right\}^{N\times 1}$. Then, $\mathbf{X}$ is fed to the node-type encoder of CircuitGPS to learn embeddings $\texttt{Embed}(\mathbf{X})\in \mathbb{R}^{N \times d_0}$. The model also considers edge attributes, where another edge-type encoder generates edge embeddings $\texttt{Embed}(\mathbf{E}) \in \mathbb{R}^{N_E \times d_0}$. 

The aforementioned attributes are critical for pre-training a link prediction model. However, for the downstream regression, the coupling capacitance is also related to other information, such as the geometric size of devices and the number of devices connecting to a net. Thus, we collect more detailed statistics for each type of node, including the design parameters in AMS circuits. We pad these statistics into another node attribute matrix $\mathbf{X}_{C} \in \mathbb{R}^{N\times d_C}$ that is input into the task-specific head of CircuitGPS. Table \ref{tab:xc} lists the definition of each dimension in $\mathbf{X}_{C}$ for different node types.

\begin{table}[tb]
\setlength{\abovecaptionskip}{0pt}
\setlength{\belowcaptionskip}{0pt}
\caption{Dimension Definition in $\mathbf{X}_{C}$.}\label{tab:xc}
\begin{center}
    \begin{tabular}{lll}
    \toprule
    Node Type & Dim. & Definition  \\ \midrule
    \multirow{13}{*}{\makecell[l]{$x_i=0$ \\ for net node}} 
    & 0 & \# of connected transistors  \\
    & 1 & \# of connected gate terminals \\
    & 2 & \# of connected source/drain terminals \\
    & 3 & \# of connected base terminals \\
    & 4 & Total width of connected transistor \\
    & 5 & Total length of connected transistor \\
    & 6 & \# of connected capacitors \\
    & 7 & Total length of connected capacitors \\
    & 8 & Total \# of connected capacitor fingers \\
    & 9 & \# of connected resistors \\
    & 10 & Total width of connected resistors \\
    & 11 & Total length of connected resistors \\
    & 12 & \# of connected ports \\ \midrule
    \multirow{11}{*}{\makecell[l]{$x_i=1$ \\ for device node }} 
    & 0 & Multiplier of transistors \\
    & 1 & Length of the transistor \\
    & 2 & Width of the transistor \\
    & 3 & Multiplier of connected resistors \\
    & 4 & Length of resistor \\
    & 5 & Width of resistor \\
    & 6 & Multiplier of connected capacitor \\
    & 7 & Length of capacitor \\
    & 8 & \# of capacitor fingers \\
    & 9 & \# of ports in the device instance \\
    & 10 & Type code of the device instance \\ \midrule
    \multirow{2}{*}{\makecell[l]{$x_i=2$ \\for pin node}} 
    & 0 & Pin types (G/D/S/B for MOS) \\ 
    &   &     \\ \bottomrule
    \end{tabular}
\end{center}
\end{table}

\subsection{Positional Encoding}\label{sec:pe}
Positional encodings (PEs) are proposed to estimate the position in space of a given node within the graph. Hence, when two nodes are close to each other within a graph, their PE should also be close. A common approach is to compute the pair-wise distance between each pair of nodes or their eigenvectors in the entire graph as proposed in \cite{ying2021graphormer, kreuzer2021rethinking}, but this is not compatible with linear Transformers (such as Performer \cite{performer}) as it requires to materialize the entire attention matrix. A better fitting solution is to use the eigenvectors of the graph Laplacian or their gradient  \cite{dwivedi2020benchmarking, beaini2021directional_dgn, kreuzer2021rethinking}. However, calculating those PEs is both time- and resource-consuming due to the large scale of subgraphs sampled from AMS circuits. For this reason, a more computationally efficient PE is demanded by our application.

The proposed PE, named double anchor shortest path distance (DSPD), is a relative PE according to the PE/SE categories proposed in \cite{rampasek2022GPS, black2024comparing}.
In each subgraph $G^{1}_{m,n}$ with anchors $m~\text{and}~n$, we calculate the shortest path distance $\left(d(i, m), d(i, n)\right)$ to the two anchors for each node $i$. The DSPD is inspired by the double-radius node labeling (DRNL) in SEAL, where the authors iteratively assign larger labels to nodes with a larger distance (radius) with respect to both anchor nodes. The labeling principle is a perfect hashing function that allows fast closed-form computations. 
Differently, the DSPD encoding does not compute a unique label; it directly stores the distance vector $\left(d(i, m), d(i, n)\right)$. The PE encoder then uses the distances to generate $\mathbf{D}_{0}$ and $\mathbf{D}_{1}$. We concatenate them with the node-type embeddings to form the final node attribute matrix,
\begin{equation}
\mathbf{X}^0=\mathbf{D}_{0} \oplus \mathbf{D}_{1} \oplus \texttt{Embed}(\mathbf{X}),
\label{eq:x}
\end{equation}
which is the input of the first GPS layer.

\begin{table}[tb]
\centering
\caption{Comparison of Different PEs in Link Prediction.}\label{tab:pecomp}
\begin{tabular}{l|llll} \toprule
PE & Acc. & F1 & AUC & Time/G (s) \\ \midrule
w/o PE & 0.8867 & 0.9120 & 0.9393 & N/A \\
$\mathbf{X}_C$ & 0.9066 & 0.9261 & 0.9629 & N/A \\
DRNL & 0.9505 & 0.9640 & 0.9698 & \textbf{0.0170} \\
RWSE & 0.8931 & 0.9255 & 0.8612 & 0.1296 \\
LapPE & 0.9561 & 0.9680 & 0.9697 & 0.1934 \\ 
DSPD & \textbf{0.9618} & \textbf{0.9720} & \textbf{0.9774} & 0.0173 \\ \bottomrule
\end{tabular}
\end{table}

Table \ref{tab:pecomp} shows the prediction accuracy of CircuitGPS on link prediction with different PEs, including circuit statistic matrix $\mathbf{X}_C$, RWSE proposed by \cite{dwivedi2022LPE}, LapPE used by \cite{dwivedi2020generalization,kreuzer2021rethinking}, DRNL proposed by SEAL, and our DSPD. We take the AMS design \textsf{\footnotesize{SSRAM}} \cite{tscache} as a small training dataset and the circuit \textsf{\footnotesize{DIGITAL\_CLK\_GEN}}~as a test dataset to evaluate CircuitGPS in a zero-shot learning setting. The proposed DSPD demonstrates the highest accuracy while ensuring computational efficiency.

\newtheorem{observation}{Observation}
\begin{observation} \label{obs:1}
Taking the circuit statistic matrix $\mathbf{X}_C$ as a positional encoding reduces the model performance compared to employing other PEs.
\end{observation}

This counterintuitive phenomenon was also reported by Shi et al.~\cite{shi2024structural}. In their work, removing original node features of neighbor nodes led to slight performance gains for SEAL on multiple link prediction datasets. While seemingly paradoxical, this result is explicable: in link prediction tasks, neighbor node features may introduce task-irrelevant noise. For instance, in our context, the specific device types connected to a target net are irrelevant for predicting coupling effects between nodes. This critical observation motivated our architectural redesign, transitioning from GraphGPS to the simplified CircuitGPS framework.  

\subsection{Hybrid Graph Transformer}\label{sec:hgt}

\newcommand{\mpnn}{\texttt{MPNN}}
\newcommand{\attn}{\texttt{GlobalAttn}}
\newcommand{\X}{\mathbf{X}}
\newcommand{\E}{\mathbf{E}}
\newcommand{\A}{\mathbf{A}}

The hybrid GT model used in this work is built from a general, powerful, and scalable architecture named GraphGPS \cite{rampasek2022GPS}. Each GPS layer is a parallel MPNN+Transformer layer, as shown in Fig. \ref{fig:framework}. 
The update function is
\begin{eqnarray}
    \X^{\ell+1}, \E^{\ell+1} &=& \texttt{GPS}^{\ell} \left( \X^{\ell}, \E^{\ell}, \A \right), 
\end{eqnarray}
and the hidden features are computed as

\begin{eqnarray}
    \X^{\ell+1}_M, \ \E^{\ell+1} &=& \mpnn_e^{\ell} \left(\X^{\ell}, \E^{\ell}, \A \right),\\
    \X^{\ell+1}_A
    &=&  \attn^{\ell} \left(\X^{\ell}  \right),\\
    \X^{\ell+1} &=&
    \texttt{MLP}^{\ell}\left(\X^{\ell+1}_M + \X^{\ell+1}_A\right),
    \label{eqn:layer_equation}
\end{eqnarray}
where $\A \in \mathbb{R}^{N \times N}$ is the adjacency matrix; $\X^{\ell} \in \mathbb{R}^{N \times d_\ell}, \E^{\ell} \in \mathbb{R}^{E \times d_\ell}$ are the $d_\ell$-dimensional node and edge attributes, respectively; $\mpnn_e^{\ell}$ and $\attn^{\ell}$ are instances of an MPNN with edge attributes and a global attention mechanism, respectively; $\texttt{MLP}^{\ell}$ is a 2-layer MLP block. After each functional block (an \mpnn~layer, a
\attn~layer, or an $\texttt{MLP}$), we apply residual connections followed by batch normalization (BN), which are omitted in \eqref{eqn:layer_equation} for clarity.
Edge attributes are only passed to the \mpnn~layer. Due to the scalability of GPS, both \mpnn~and \attn~layers are modular, where \mpnn~can be any function that acts on a local neighborhood and \attn~can be any fully connected layer.

Based on Observation \ref{obs:1}, we customize the GraphGPS with a task-specific head. We add two linear projection layers and an embedding layer in the head to project circuit statistics $\mathbf{X}_C$ (Table \ref{tab:xc}) into a unified attribute matrix $\mathbf{C} \in \mathbb{R}^{N\times d_L}$. For a node $i$ in the subgraph with node type $x_i$ and circuit statistic $x_C^i$, $c_i$ is produced by
\begin{eqnarray}
c_i = 
\begin{cases}
\texttt{Linear}(x_C^i),~ &\text{if}~ x_i=0; \\ 
\texttt{Linear}(x_C^i),~ &\text{if}~ x_i=1; \\ 
\texttt{Embed}(x_C^i),~ &\text{if}~ x_i=2.
\end{cases}
\end{eqnarray}
Finally, the hidden input of the task-specific head \texttt{MLP} is
\begin{eqnarray}
\mathbf{X}_H = \texttt{Pool}(\mathbf{X}^L+\mathbf{C}).
\end{eqnarray}

To find a better layer architecture, we produced an ablation study of different layer types on a small training set as shown in Table \ref{tab:ablink}. The alternatives are a global attention-based Transformer, a linear Transformer (Performer \cite{performer}), a pure MPNN (GatedGCN \cite{gatedgcn}), and their combinations. 

\begin{observation} \label{obs:2}
Classic MPNN models achieve strong performance, matching or even exceeding that of pure Transformers or MPNN+Transformers. 
\end{observation}

This observation coincides with the results reported by Luo et al. in their recent work \cite{luo2024classic}. According to their findings, this strong performance is driven by normalization, dropout, and residual connections. All these features have been integrated into CircuitGPS and can be adjusted with the hyperparameters in configuration files. Beyond excellent performance, MPNN requires lower training overhead and converges more easily. 

\begin{table}[tb]
\setlength\tabcolsep{2.5pt}
\centering
\caption{Ablation Study of Different GPS Layer Configuration on Link Prediction (\textbf{best}, \underline{second best}).}\label{tab:ablink}
\resizebox{\linewidth}{!}{
    \begin{tabular}{ll|lllll} \toprule
    MPNN & Attention & Acc. & F1 & AUC & Time(s) & \#Param. \\ \midrule
    None & Performer & 0.9458 & 0.9602 & 0.9668 & 1663.0 & 762,390 \\ 
    None & Transformer & 0.9456 & 0.9601 & 0.9187 & 3490.0 & 778,833 \\
    GatedGCN & Performer& 0.9618 & 0.9720 & 0.9774 & \underline{1446.1} & 752,785 \\
    GatedGCN & Transformer & \textbf{0.97011} & \textbf{0.9780} & \textbf{0.9980} & 2832.9 & 540,337 \\ 
    GatedGCN & None & \underline{0.9693} & \underline{0.9775} & \underline{0.9848} & \textbf{965.6} & 724,854 \\
    \bottomrule
    \end{tabular}
}
\end{table}

\subsection{Model Fine-tuning}
Models trained with few-shot learning techniques are often better at generalizing to new tasks or classes, as they focus on capturing transferable knowledge rather than memorizing specific data patterns. 
The pre-trained CircuitGPS on link prediction aims to learn an informational representation of subgraphs. Fine-tuning can be implemented in two ways: (1) freezing the encoders and GPS layers while training only the task-specific head, which significantly accelerates the convergence, or (2) continuing to train all parameters in CircuitGPS using the pre-trained ones as initialization, further improving edge regression accuracy (Table \ref{tab:edgereg}).



\section{Experiments}\label{sec:experim}

Our implementation is based on PyG \cite{FeyLenssen2019PyG}, GraphGym \cite{you2020design}, and GraphGPS \cite{rampasek2022GPS} modules. All experiments were run in a shared computing cluster. The cluster has 40 Intel Xeon Silver CPUs with 128GB of memory and four Nvidia 4090 (24GB) GPUs. Each experiment was executed using four to six CPUs, a single GPU, and 128 GB of RAM.

Circuit statistics were obtained by converting the netlist to a graph. 
Before the enclosing subgraph sampling, we followed the setup of SEAL \cite{seal}, where both the positive edges and the negative edges were injected into the original circuit graph. 
The sampling procedure was implemented with the multiprocessing package to reduce the sampling time overhead. 
We prepared the full schematic netlists to extract the graph and features, and the post-layout netlists (SPF files) to collect the ground-truth labels and targets. The configuration files for experiments in this section can be found in our \href{https://github.com/ShenShan123/CirGPS.git}{Git repository}.

\vspace{-2pt}
\subsection{Dataset}\label{dataset}

\begin{table}[!b]
\setlength\tabcolsep{2.5pt}
\centering
\caption{AMS Circuit Dataset Statistics. 
}\label{tab:dataset}
\resizebox{\linewidth}{!}{
    \begin{tabular}{l|llllll}
    \toprule
    Split & Dataset & $N$ & $N_E$ & \#Links & $N/G^1_{m,n}$ & $N_E/G^1_{m,n}$ \\ \midrule
    \multirow{3}{*}{Train} 
    & \ssram & 87K & 134K & 131K & 153 & 917 \\
    & \ultra & 3.5M & 13.4M & 166K & 257 & 1,476\\
    & \sandwich & 4.3M & 13.3M & 154K & 472 & 2,540 \\ \midrule
    \multirow{3}{*}{Test} 
    & \digtime & 17K & 36K & 4K & 417 & 2,403 \\
    & \timectrl & 18K & 44K & 5K & 59 & 387 \\
    & \sarray & 144K & 352K & 110K & 150 & 803 \\ \bottomrule
    \end{tabular}
}
\end{table}

Table \ref{tab:dataset} summarizes the AMS designs used as datasets in this work. All designs are under 28nm CMOS technology. \textsf{\footnotesize{SSRAM}} \cite{tscache} is a small design with high energy efficiency, consisting of standard digital cells and static random access memory (SRAM) arrays. 
\textsf{\footnotesize{ULTRA8T}} SRAM \cite{shen2024ultra8t} is a multi-voltage design that contains large analog modules and SRAM arrays.
\textsf{\footnotesize{SANDWICH-RAM}} \cite{cim1} is evenly divided between digital circuits for computing and SRAM arrays for storage, forming a sandwich-like structure.  
Due to the large scale of the datasets, training with all coupling links results in unacceptable time overhead. The redundancy of capacitance also leads to inefficient training \cite{shen2024deep}. Thus, we only sample a fraction of the links to reduce the training time per epoch, as shown in the ``\#links'' column of Table \ref{tab:dataset}.

In terms of test data sets, \textsf{\footnotesize{DIGITAL\_CLK\_GEN}} is composed of digital cells and SRAM columns and is used to generate the internal clock in SRAM. It is the most challenging test case of our experiments. \textsf{\footnotesize{TIMING\_CONTROL}} consists of standard digital cells to produce control signals for the SRAM module. \textsf{\footnotesize{ARRAY\_128\_32}} is a single 128-row 32-column SRAM array. None of the test datasets are part of the training datasets, and we ensure that all test data is not visible during training and validation. Thus, all results in this work are reported from zero-shot learning.

\subsection{Link Prediction Task}\label{sec:linkpred}
In this subsection, we show the prediction performance of CircuitGPS on the link prediction task.
We adapted the two existing MPNN-based models, ParaGraph \cite{ParaGraph} and DLPL-Cap \cite{shen2024deep}, to our coupling prediction task for comparison purposes. These two models didn't involve any sampling technique or PE, and they directly used the entire graphs and the circuit statistic matrix $\mathbf{X}_C$ as input. We tuned the model hyperparameters to improve their performance.
Table \ref{tab:linkcomp} lists the performance comparison. After combining enclosing subgraph sampling and DSPD encoding, CircuitGPS has demonstrated a huge improvement in terms of all performance metrics, improving accuracy by at least 20\%.

\begin{table}[!t]
\centering
\caption{Accuracy Comparison of Different  Methods on Link Prediction.}\label{tab:linkcomp}
\setlength\tabcolsep{1.5pt}
\resizebox{\linewidth}{!}{
\begin{tabular}{l|lll|lll|lll}
\toprule
Dataset & \multicolumn{3}{c|}{\digtime} & \multicolumn{3}{c|}{\timectrl} & \multicolumn{3}{c}{\sarray} \\ 
Metrics & Acc. & F1 & AUC & Acc. & F1 & AUC & Acc. & F1 & AUC \\ \midrule
ParaGraph & 0.768 & 0.847 & 0.870 & 0.754 & 0.841 & 0.865 & 0.720 & 0.776 & 0.823 \\
DLPL-Cap & 0.761 & 0.841 & 0.864 & 0.750 & 0.839 & 0.865 & 0.756 & 0.832 & 0.825 \\
CircuitGPS & \textbf{0.972} & \textbf{0.979} & \textbf{0.992} & \textbf{0.989} & \textbf{0.992} & \textbf{0.998} & \textbf{0.980} & \textbf{0.985} & \textbf{0.999}  \\ \bottomrule
\end{tabular}
}
\end{table}

\subsection{Edge Regression Task}\label{sec:edgereg}

\begin{table}[b]
\centering
\caption{Error Comparison of Different  Methods on Edge Regression.} 
\label{tab:edgereg}
\setlength\tabcolsep{1.5pt}
\resizebox{\linewidth}{!}{
    \begin{tabular}{@{}l|lll|lll|lll@{}} \toprule
    Dataset & \multicolumn{3}{c|}{\digtime} & \multicolumn{3}{c|}{\timectrl} & \multicolumn{3}{c}{\sarray} \\ 
    Metrics & MAE$\downarrow$ & RMSE$\downarrow$ & \multicolumn{1}{c|}{$R^2$} & MAE$\downarrow$ & RMSE$\downarrow$ & \multicolumn{1}{c|}{$R^2$} & MAE$\downarrow$ & RMSE$\downarrow$ & \multicolumn{1}{c}{$R^2$} \\ \midrule
    ParaGraph & 0.153 & 0.212 & 0.470 & 0.154 & 0.214 & 0.590 & 0.181 & 0.260 & 0.211 \\
    DLPL-Cap & 0.160 & 0.223 & 0.414 & 0.157 & 0.217 & 0.579 & 0.176 & 0.239 & 0.331 \\
    CircuitGPS & 0.083 & 0.130 & 0.801 & 0.043 & 0.097 & 0.915 & 0.048 & 0.120 & 0.831 \\
    CircuitGPS\textsubscript{head-ft} & 0.086 & 0.125 & 0.816 & 0.085 & 0.131 & 0.847 & 0.075 & 0.120 & 0.831 \\
    CircuitGPS\textsubscript{all-ft} & \textbf{0.072} & \textbf{0.120} & \textbf{0.833} & \textbf{0.042} & \textbf{0.093} & \textbf{0.923} & \textbf{0.040} & \textbf{0.074} & \textbf{0.936} \\
    \bottomrule
    \end{tabular}
}
\end{table}

The edge regression task is to predict the exact coupling capacitance of the given node pair. The negative links injected during pre-training were assigned zero capacitance. As capacitance values spanned several magnitudes, we only kept the nets with the capacitance value $1^{-21}F\leq \boldsymbol y_{C} \leq 1^{-15}F$. Then we normalized the statistic attribute matrix $\mathbf{X_C}$ and the capacitance values to $[0, 1]$ to avoid numerical instability. As Table \ref{tab:edgereg} lists, CircuitGPS reduces the MAE of capacitance estimation by at least 0.067 compared to other methods.

We also conducted an ablation study of different GPS layer configurations for edge regression on the small training set \textsf{\footnotesize{SSRAM}} and the test dataset \textsf{\footnotesize{DIGITAL\_CLK\_GEN}}. Table \ref{tab:abreg} lists the results. Again, the classic GatedGCN performs strongly, demonstrating Observation \ref{obs:2}. Pure Transformer layers have the worst performance due to the limited number of parameters that our computing platform can handle.

\begin{table}[htb]
\setlength\tabcolsep{2.5pt}
\centering
\caption{Ablation Study of Different GPS Layer Configuration on Edge Regression.}\label{tab:abreg}
\resizebox{\linewidth}{!}{
    \begin{tabular}{ll|lllll} \toprule
    MPNN & Attention & MAE$\downarrow$ & RMSE$\downarrow$ & \multicolumn{1}{c}{$R^2$} & Time(s) &\#Param. \\ \midrule
    None & Performer & 0.0854 & 0.1439 & 0.7563 & 1437.3 & 736,871 \\
    None & Transformer & 0.1051 & 0.1502 & 0.7351 & 2203.6 & 480,167 \\
    GatedGCN & Performer & \textbf{0.0705} & 0.12974 & 0.8019 & 2667.9 & 751,311 \\
    GatedGCN & Transformer & 0.0772 & 0.1358 & 0.7831 & 4765.2 & 506,703 \\
    GatedGCN & None & 0.0718 & \textbf{0.1233} & \textbf{0.8212} & \textbf{931.5} & 723,380 \\ \bottomrule
    \end{tabular}
}
\end{table}

\subsection{Extending Study on Node Regression}\label{sec:ext} 
Beyond link-level tasks, CircuitGPS can also fulfill node-level tasks. We trained CircuitGPS to predict the ground parasitic capacitance on each net/pin. In this task, we didn't involve any negative link injection, and a 2-hop subgraph sampling was adopted to obtain more neighbors around the singular anchor node. The DSPD became identical distances $\mathbf{D}_0=\mathbf{D}_1$. The architecture remained the same as that in the link-level task.
Table \ref{tab:nodereg} shows the predicted errors. CircuitGPS achieves the best performance compared to the other two models in the node regression task. DLPL-Cap has the largest prediction error for the test dataset. This method first learns from a node classification task and then trains a specific regressor for each node class. Therefore, it has data-sensitive characteristics that limit generalization.

\begin{table}[tb]
\setlength\tabcolsep{1.5pt}
\centering
\caption{Error Comparison of Different  Methods on Node Regression.}\label{tab:nodereg}
\resizebox{\linewidth}{!}{
    \begin{tabular}{l|lll|lll|lll} \toprule
    Dataset & \multicolumn{3}{c|}{\digtime} & \multicolumn{3}{c|}{\timectrl} & \multicolumn{3}{c}{\sarray} \\ 
    Metrics & MAE$\downarrow$ & RMSE$\downarrow$ & \multicolumn{1}{c|}{$R^2$} & MAE$\downarrow$ & RMSE$\downarrow$ & \multicolumn{1}{c|}{$R^2$} & MAE$\downarrow$ & RMSE$\downarrow$ & \multicolumn{1}{c}{$R^2$} \\ \midrule
    ParaGraph & 0.101 & 0.144 & 0.313 & 0.112 & 0.154 & 0.462 & 0.114 & 0.174 & 0.002 \\
    DLPL-Cap & 0.137 & 0.208 & 0.364 & 0.096 & 0.137 & 0.379 & 0.097 & 0.136 & 0.390 \\
    CircuitGPS & \textbf{0.072} & \textbf{0.104} & \textbf{0.643} & \textbf{0.088} & \textbf{0.132} & \textbf{0.602} & \textbf{0.078} & \textbf{0.101} & \textbf{0.637}  \\
    \bottomrule
    \end{tabular}
}
\end{table}

\begin{figure}[tb]
    \setlength{\abovecaptionskip}{0pt}
    \setlength{\belowcaptionskip}{0pt}
    \centering
    \includegraphics[width=0.8\linewidth]{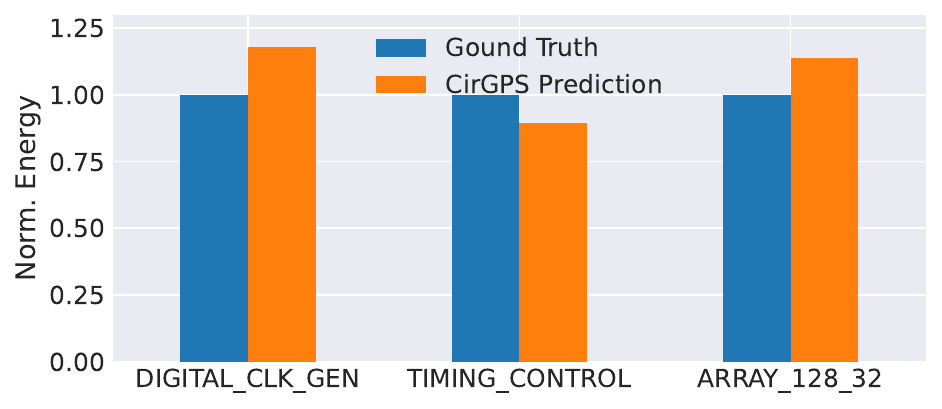}
    \caption{The simulated energy consumption with ground truth parasitic capacitance vs. predictions of CircuitGPS.}
    \label{fig:energy}
\end{figure}

Finally, we conducted the circuit simulations using SPICE to validate the predicted capacitance of CircuitGPS (without involving any parasitic resistance). Simulated energy consumption can further demonstrate the effectiveness of CircuitGPS, as this metric is primarily determined by parasitic capacitance in AMS circuits. In Fig. \ref{fig:energy}, the mean absolute percentage error of three test cases is 14.5\%.

\section{Conclusion}\label{conclu}
This work introduces CircuitGPS, a few-shot learning framework designed to estimate parasitic effects in AMS circuits. The method incorporates subgraph sampling, a computationally efficient PE, and a customized GT. The robust generalization and transferability of CircuitGPS make it adaptable to a variety of tasks, including link prediction (for coupling effect analysis), edge regression (to estimate coupling capacitance), and node regression (for ground capacitance estimation).

\bibliographystyle{IEEEtran}
\bibliography{refs}

\begin{thebibliography}{10}
\providecommand{\url}[1]{#1}
\csname url@samestyle\endcsname
\providecommand{\newblock}{\relax}
\providecommand{\bibinfo}[2]{#2}
\providecommand{\BIBentrySTDinterwordspacing}{\spaceskip=0pt\relax}
\providecommand{\BIBentryALTinterwordstretchfactor}{4}
\providecommand{\BIBentryALTinterwordspacing}{\spaceskip=\fontdimen2\font plus
\BIBentryALTinterwordstretchfactor\fontdimen3\font minus \fontdimen4\font\relax}
\providecommand{\BIBforeignlanguage}[2]{{%
\expandafter\ifx\csname l@#1\endcsname\relax
\typeout{** WARNING: IEEEtran.bst: No hyphenation pattern has been}%
\typeout{** loaded for the language `#1'. Using the pattern for}%
\typeout{** the default language instead.}%
\else
\language=\csname l@#1\endcsname
\fi
#2}}
\providecommand{\BIBdecl}{\relax}
\BIBdecl

\bibitem{wang2020generalizing}
Y.~Wang, Q.~Yao, J.~T. Kwok, and L.~M. Ni, ``Generalizing from a few examples: A survey on few-shot learning,'' \emph{ACM computing surveys (csur)}, vol.~53, no.~3, pp. 1--34, 2020.

\bibitem{hakhamaneshi2022pretraining}
K.~Hakhamaneshi, M.~Nassar, M.~Phielipp, P.~Abbeel, and V.~Stojanovic, ``Pretraining graph neural networks for few-shot analog circuit modeling and design,'' \emph{IEEE Transactions on Computer-Aided Design of Integrated Circuits and Systems}, vol.~42, no.~7, pp. 2163--2173, 2022.

\bibitem{yu2014advanced}
W.~Yu and X.~Wang, \emph{Advanced field-solver techniques for RC extraction of integrated circuits}.\hskip 1em plus 0.5em minus 0.4em\relax Springer, 2014.

\bibitem{yu2009variational}
W.~Yu, C.~Hu, and W.~Zhang, ``Variational capacitance extraction of on-chip interconnects based on continuous surface model,'' in \emph{Proceedings of the 46th Annual Design Automation Conference}, 2009, pp. 758--763.

\bibitem{seal}
M.~Zhang and Y.~Chen, ``Link prediction based on graph neural networks,'' \emph{Advances in neural information processing systems}, vol.~31, 2018.

\bibitem{rampasek2022GPS}
L.~Ramp\'{a}\v{s}ek, M.~Galkin, V.~P. Dwivedi, A.~T. Luu, G.~Wolf, and D.~Beaini, ``{Recipe for a General, Powerful, Scalable Graph Transformer},'' \emph{Advances in Neural Information Processing Systems}, vol.~35, 2022.

\bibitem{ravi2017optimization}
S.~Ravi and H.~Larochelle, ``Optimization as a model for few-shot learning,'' in \emph{International conference on learning representations}, 2017.

\bibitem{zhang2017weisfeiler}
M.~Zhang and Y.~Chen, ``Weisfeiler-lehman neural machine for link prediction,'' in \emph{Proceedings of the 23rd ACM SIGKDD international conference on knowledge discovery and data mining}, 2017, pp. 575--583.

\bibitem{lu2011link}
L.~L{\"u} and T.~Zhou, ``Link prediction in complex networks: A survey,'' \emph{Physica A: statistical mechanics and its applications}, vol. 390, no.~6, pp. 1150--1170, 2011.

\bibitem{kalyan2021ammus_transformer_survey}
K.~S. Kalyan, A.~Rajasekharan, and S.~Sangeetha, ``Ammus: A survey of transformer-based pretrained models in natural language processing,'' \emph{arXiv:2108.05542}, 2021.

\bibitem{han2022survey_vision_transformer}
K.~Han, Y.~Wang, H.~Chen, X.~Chen, J.~Guo, Z.~Liu, Y.~Tang, A.~Xiao, C.~Xu, Y.~Xu \emph{et~al.}, ``A survey on vision transformer,'' \emph{IEEE Transactions on Pattern Analysis and Machine Intelligence}, 2022.

\bibitem{dwivedi2020generalization}
V.~P. Dwivedi and X.~Bresson, ``A generalization of transformer networks to graphs,'' \emph{arXiv:2012.09699}, 2020.

\bibitem{kreuzer2021rethinking}
D.~Kreuzer, D.~Beaini, W.~L. Hamilton, V.~L{\'e}tourneau, and P.~Tossou, ``Rethinking graph transformers with spectral attention,'' in \emph{Advances in Neural Information Processing Systems}, 2021.

\bibitem{ying2021graphormer}
C.~Ying, T.~Cai, S.~Luo, S.~Zheng, G.~Ke, D.~He, Y.~Shen, and T.-Y. Liu, ``Do transformers really perform badly for graph representation?'' in \emph{Advances in Neural Information Processing Systems}, 2021.

\bibitem{dwivedi2022LPE}
V.~P. Dwivedi, A.~T. Luu, T.~Laurent, Y.~Bengio, and X.~Bresson, ``Graph neural networks with learnable structural and positional representations,'' in \emph{International Conference on Learning Representations}, 2022.

\bibitem{beaini2021directional_dgn}
D.~Beaini, S.~Passaro, V.~L{\'e}tourneau, W.~Hamilton, G.~Corso, and P.~Li{\`o}, ``Directional graph networks,'' in \emph{International Conference on Machine Learning}.\hskip 1em plus 0.5em minus 0.4em\relax PMLR, 2021, pp. 748--758.

\bibitem{topping2021understanding_ricci}
J.~Topping, F.~Di~Giovanni, B.~P. Chamberlain, X.~Dong, and M.~M. Bronstein, ``Understanding over-squashing and bottlenecks on graphs via curvature,'' \emph{arXiv:2111.14522}, 2021.

\bibitem{ParaGraph}
H.~Ren, G.~F. Kokai, W.~J. Turner, and T.-S. Ku, ``{ParaGraph}: Layout parasitics and device parameter prediction using graph neural networks,'' in \emph{Proc. DAC}, 2020, pp. 1--6.

\bibitem{shen2024deep}
S.~Shen, D.~Yang, Y.~Xie, C.~Pei, B.~Yu, and W.~Yu, ``Deep-learning-based pre-layout parasitic capacitance prediction on sram designs,'' in \emph{Proceedings of the Great Lakes Symposium on VLSI 2024}, 2024, pp. 440--445.

\bibitem{performer}
K.~M. Choromanski, V.~Likhosherstov, D.~Dohan, X.~Song, A.~Gane, T.~Sarlos, P.~Hawkins, J.~Q. Davis, A.~Mohiuddin, L.~Kaiser \emph{et~al.}, ``Rethinking attention with performers,'' in \emph{International Conference on Learning Representations}, 2020.

\bibitem{dwivedi2020benchmarking}
V.~P. Dwivedi, C.~K. Joshi, T.~Laurent, Y.~Bengio, and X.~Bresson, ``Benchmarking graph neural networks,'' \emph{arXiv:2003.00982}, 2020.

\bibitem{black2024comparing}
M.~Black, Z.~Wan, G.~Mishne, A.~Nayyeri, and Y.~Wang, ``Comparing graph transformers via positional encodings,'' \emph{arXiv preprint arXiv:2402.14202}, 2024.

\bibitem{tscache}
S.~Shen, T.~Shao, X.~Shang, Y.~Guo, M.~Ling, J.~Yang, and L.~Shi, ``{TS} cache: A fast cache with timing-speculation mechanism under low supply voltages,'' \emph{IEEE Transactions on Very Large Scale Integration (VLSI) Systems}, vol.~28, no.~1, pp. 252--262, 2019.

\bibitem{shi2024structural}
L.~Shi, B.~Hu, D.~Zhao, J.~He, Z.~Zhang, and J.~Zhou, ``Structural information enhanced graph representation for link prediction,'' in \emph{Proceedings of the AAAI Conference on Artificial Intelligence}, vol.~38, no.~13, 2024, pp. 14\,964--14\,972.

\bibitem{gatedgcn}
X.~Bresson and T.~Laurent, ``Residual gated graph convnets,'' \emph{arXiv preprint arXiv:1711.07553}, 2017.

\bibitem{luo2024classic}
Y.~Luo, L.~Shi, and X.-M. Wu, ``Classic gnns are strong baselines: Reassessing gnns for node classification,'' \emph{arXiv preprint arXiv:2406.08993}, 2024.

\bibitem{FeyLenssen2019PyG}
M.~Fey and J.~E. Lenssen, ``Fast graph representation learning with {PyTorch Geometric},'' in \emph{ICLR Workshop on Representation Learning on Graphs and Manifolds}, 2019.

\bibitem{you2020design}
J.~You, R.~Ying, and J.~Leskovec, ``Design space for graph neural networks,'' in \emph{Advances in Neural Information Processing Systems}, 2020.

\bibitem{shen2024ultra8t}
S.~Shen, H.~Xu, Y.~Zhou, M.~Ling, and W.~Yu, ``Ultra8t: A sub-threshold 8t sram with leakage detection,'' \emph{Integration}, vol.~98, p. 102233, 2024.

\bibitem{cim1}
J.~Yang, Y.~Kong, Z.~Wang, Y.~Liu, B.~Wang, S.~Yin, and L.~Shi, ``{24.4 sandwich-RAM: An energy-efficient in-memory BWN architecture with pulse-width modulation},'' in \emph{Proc. Int. Solid-State Circuits Conf. (ISSCC)}, 2019, pp. 394--396.

\end{thebibliography}

\end{document}